\documentclass[11pt]{article}
\usepackage{acl2016}
\usepackage{times}
\usepackage{latexsym}

\usepackage{amsmath}
\usepackage{amsfonts}
\usepackage{mathtools}
\usepackage{bm}

\usepackage{booktabs,array}
\newcolumntype{C}[1]{>{\centering\arraybackslash\hspace{0pt}}p{#1}}
\newcolumntype{L}[1]{>{\raggedright\arraybackslash}p{#1}}

\usepackage[pdftex]{graphicx}
\graphicspath{{figures/}}

\usepackage{enumitem}
\setlist{parsep=0pt,listparindent=\parindent}

\usepackage{tikz}

\aclfinalcopy 
\def\aclpaperid{47} 

\title{A Comparative Study on Vocabulary Reduction\\for Phrase Table Smoothing}

\author{Yunsu Kim, Andreas Guta, Joern Wuebker$^*$, \textnormal{and} Hermann Ney\\
	    Human Language Technology and Pattern Recognition Group\\
	    RWTH Aachen University, Aachen, Germany\\
      {\tt \{surname\}@cs.rwth-aachen.de}\\
      $^*$Lilt, Inc.\\
	    {\tt joern@lilt.com}}

\date{}

\begin{document}
\maketitle
\begin{abstract}
This work systematically analyzes the smoothing effect of vocabulary reduction for phrase translation models. We extensively compare various word-level vocabularies to show that the performance of smoothing is not significantly affected by the choice of vocabulary. This result provides empirical evidence that the standard phrase translation model is extremely sparse. Our experiments also reveal that vocabulary reduction is more effective for smoothing large-scale phrase tables.
\end{abstract}

\section{Introduction}
Phrase-based systems for statistical machine translation (SMT) \cite{Zens02,Koehn03} have shown state-of-the-art performance over the last decade. However, due to the huge size of phrase vocabulary, it is difficult to collect robust statistics for lots of phrase pairs. The standard phrase translation model thus tends to be sparse \cite{Koehn10}.

A fundamental solution to a sparsity problem in natural language processing is to reduce the vocabulary size. By mapping words onto a smaller label space, the models can be trained to have denser distributions \cite{Brown92,Miller04,Koo08}. Examples of such labels are part-of-speech (POS) tags or lemmas.

In this work, we investigate the vocabulary reduction for phrase translation models with respect to various vocabulary choice. We evaluate two types of smoothing models for phrase translation probability using different kinds of word-level labels. In particular, we use automatically generated word classes \cite{Brown92} to obtain label vocabularies with arbitrary sizes and structures. Our experiments reveal that the vocabulary of the smoothing model has no significant effect on the end-to-end translation quality. For example, a randomized label space also leads to a decent improvement of \textsc{Bleu} or \textsc{Ter} scores by the presented smoothing models.

We also test vocabulary reduction in translation scenarios of different scales, showing that the smoothing works better with more parallel corpora.

\section{Related Work}
\label{sec:related}
\newcite{Koehn07} propose integrating a label vocabulary as a factor into the phrase-based SMT pipeline, which consists of the following three steps: mapping from words to labels, label-to-label translation, and generation of words from labels. \newcite{Rishoj11} verify the effectiveness of word classes as factors. Assuming probabilistic mappings between words and labels, the factorization implies a combinatorial expansion of the phrase table with regard to different vocabularies.

\newcite{Wuebker13} show a simplified case of the factored translation by adopting hard assignment from words to labels. In the end, they train the existing translation, language, and reordering models on word classes to build the corresponding smoothing models.

Other types of features are also trained on word-level labels, e.g. hierarchical reordering features \cite{Cherry13}, an $n$-gram-based translation model \cite{Durrani14}, and sparse word pair features \cite{Haddow15}. The first and the third are trained with a large-scale discriminative training algorithm.

For all usages of word-level labels in SMT, a common and important question is which label vocabulary maximizes the translation quality. \newcite{Bisazza14} compare class-based language models with diverse kinds of labels in terms of their performance in translation into morphologically rich languages. To the best of our knowledge, there is no published work on systematic comparison between different label vocabularies, model forms, and training data size for smoothing phrase translation models---the most basic component in state-of-the-art SMT systems. Our work fulfills these needs with extensive translation experiments (Section \ref{sec:exp}) and quantitative analysis (Section \ref{sec:analysis}) in a standard phrase-based SMT framework.

\section{Word Classes}
\label{sec:wc}
In this work, we mainly use unsupervised word classes by \newcite{Brown92} as the reduced vocabulary. This section briefly reviews the principle and properties of word classes.

A word-class mapping $c$ is estimated by a clustering algorithm that maximizes the following objective \cite{Brown92}:
\begin{align}
  \mathcal{L}:=\sum_{e_1^I}\sum_{i=1}^I p(c(e_i)|c(e_{i-1}))\cdot p(e_i|c(e_i))
\end{align}
for a given monolingual corpus $\{e_1^I\}$, where each $e_1^I$ is a sentence of length $I$ in the corpus. The objective guides $c$ to prefer certain collocations of class sequences, e.g. an auxiliary verb class should succeed a class of pronouns or person names. Consequently, the resulting $c\,$ groups words according to their syntactic or semantic similarity.

Word classes have a big advantage for our comparative study: The structure and size of the class vocabulary can be arbitrarily adjusted by the clustering parameters. This makes it possible to prepare easily an abundant set of label vocabularies that differ in linguistic coherence and degree of generalization.

\section{Smoothing Models}
\label{sec:wcs}
In the standard phrase translation model, the translation probability for each segmented phrase pair $(\tilde{f},\tilde{e})$ is estimated by relative frequencies:
\begin{align}
p_\text{std}(\tilde{f}|\tilde{e})=\frac{N(\tilde{f},\tilde{e})}{N(\tilde{e})}
\label{eq:std}
\end{align}
where $N$ is the count of a phrase or a phrase pair in the training data. These counts are very low for many phrases due to a limited amount of bilingual training data.

Using a smaller vocabulary, we can aggregate the low counts and make the distribution smoother. We now define two types of smoothing models for Equation \ref{eq:std} using a general word-label mapping $c$.

\subsection{Mapping All Words at Once (map-all)}
\label{sec:wctm}
For the phrase translation model, the simplest formulation of vocabulary reduction is obtained by replacing all words in the source and target phrases with the corresponding labels in a smaller space. Namely, we employ the following probability instead of Equation \ref{eq:std}:
\begin{align}
p_\text{all}(\tilde{f}|\tilde{e}) &= \frac{N(c(\tilde{f}),c(\tilde{e}))}{N(c(\tilde{e}))}
\end{align}
which we call \emph{map-all}. This model resembles the word class translation model of \newcite{Wuebker13} except that we allow any kind of word-level labels.

This model generalizes all words of a phrase without distinction between them. Also, the same formulation is applied to word-based lexicon models.

\subsection{Mapping Each Word at a Time (map-each)}
\label{sec:pcs}
More elaborate smoothing can be achieved by generalizing only a sub-part of the phrase pair. The idea is to replace one source word at a time with its respective label. For each source position $j$, we also replace the target words aligned to the source word $f_{j}$. For this purpose, we let $a_{j}\subseteq\{1, ..., |\tilde{e}|\}$ denote a set of target positions aligned to $j$. The resulting model takes a weighted average of the redefined translation probabilities over all source positions of $\tilde{f}$:
\begin{align}
\label{eq:each}
p_\text{each}(\tilde{f}|\tilde{e}) &= \sum_{j=1}^{|\tilde{f}|}w_j \cdot \frac{N(c^{(j)}(\tilde{f}),c^{(a_j)}(\tilde{e}))}{N(c^{(a_j)}(\tilde{e}))}
\end{align}
where the superscripts of $c$ indicate the positions that are mapped onto the label space. $w_j$ is a weight for each source position, where $\sum_j w_j = 1$. We call this model \emph{map-each}.

We illustrate this model with a pair of three-word phrases: $\tilde{f} = [f_1,f_2,f_3]$ and $\tilde{e} = [e_1,e_2,e_3]$ (see Figure \ref{fig:csm2-ex} for the in-phrase word alignments). The map-each model score for this phrase pair is:
\begin{figure}[!ht]
\centering
\begin{tikzpicture}
  \node at (1,0) {$f_1$};
  \node at (2,0) {$f_2$};
  \node at (3,0) {$f_3$};
  \node at (0,1) {$e_1$};
  \node at (0,2) {$e_2$};
  \node at (0,3) {$e_3$};
  \node[fill,rectangle,scale=2.0] at (1,1) {};
  \node[fill,rectangle,scale=0.5] at (1,2) {};
  \node[fill,rectangle,scale=0.5] at (1,3) {};
  \node[fill,rectangle,scale=0.5] at (2,1) {};
  \node[fill,rectangle,scale=0.5] at (2,2) {};
  \node[fill,rectangle,scale=0.5] at (2,3) {};
  \node[fill,rectangle,scale=0.5] at (3,1) {};
  \node[fill,rectangle,scale=2.0] at (3,2) {};
  \node[fill,rectangle,scale=2.0] at (3,3) {};
\end{tikzpicture}
\caption{Word alignments of a pair of three-word phrases.}
\label{fig:csm2-ex}
\end{figure}
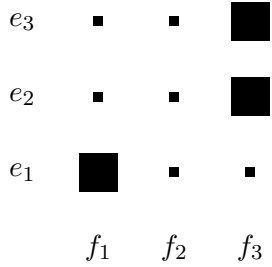
\begin{align*}
p_\text{each}(\,[&f_1,f_2,f_3]\;|\;[e_1,e_2,e_3]\,) =\\
&w_1 \cdot \frac{N([c\hspace{-0.05cm}\overbracket[0.5pt][5pt]{(f_1),f_2,f_3], [c}(e_1),e_2,e_3])}{N([c(e_1),e_2,e_3])}\\[0.3em]
+\;&w_2 \cdot \frac{N([f_1,c(f_2),f_3],[e_1,e_2,e_3])}{N([e_1,e_2,e_3])}\\
+\;&w_3 \cdot \frac{N([f_1,f_2,c\hspace{-0.05cm}\overbracket[0.5pt][5pt]{\phantom{(f_3)],[e_1,c(e_2),c}}\hspace{-2.7cm}(\overbracket[0.5pt][3pt]{f_3)],[e_1,c}(e_2),c(e_3)])}{N([e_1,c(e_2),c(e_3)])}
\end{align*}
where the alignments are depicted by line segments.

First of all, we replace $f_1$ and also $e_1$, which is aligned to $f_1$, with their corresponding labels. As $f_2$ has no alignment points, we do not replace any target word accordingly. $f_3$ triggers the class replacement of two target words at the same time. Note that the model implicitly encapsulates the alignment information.

We empirically found that the map-each model performs best with the following weight:
\begin{align}
\label{eq:fac}
w_j = \frac{N(c^{(j)}(\tilde{f}),c^{(a_j)}(\tilde{e}))}{\sum\limits_{j'=1}^{|\tilde{f}|} N(c^{(j')}(\tilde{f}),c^{(a_{j'})}(\tilde{e}))}
\end{align}
which is a normalized count of the generalized phrase pair itself. Here, the count is relatively large when $f_j$, the word to be backed off, is less frequent than other words in $\tilde{f}$. In contrast, if $f_j$ is a very frequent word and one of the other words in $\tilde{f}$ is rare, the count becomes low due to that rare word. The same logic holds for target words in $\tilde{e}$. After all, Equation \ref{eq:fac} carries more weight when a rare word is replaced with its label. The intuition is that a rare word is the main reason for unstable counts and should be backed off above all. We use this weight for all experiments in the next section.

In contrast, the map-all model merely replace all words at one time and ignore alignments within phrase pairs.

\section{Experiments}
\label{sec:exp}

\subsection{Setup}
We evaluate how much the translation quality is improved by the smoothing models in Section \ref{sec:wcs}. The two smoothing models are trained in both source-to-target and target-to-source directions, and integrated as additional features in the log-linear combination of a standard phrase-based SMT system \cite{Koehn03}. We also test linear interpolation between the standard and smoothing models, but the results are generally worse than log-linear interpolation. Note that vocabulary reduction models by themselves cannot replace the corresponding standard models, since this leads to a considerable drop in translation quality \cite{Wuebker13}.

Our baseline systems include phrase translation models in both directions, word-based lexicon models in both directions, word/phrase penalties, a distortion penalty, a hierarchical lexicalized reordering model \cite{Galley08}, a 4-gram language model, and a 7-gram word class language model \cite{Wuebker13}. The model weights are trained with minimum error rate training \cite{Och03mert}. All experiments are conducted with an open source phrase-based SMT toolkit Jane 2 \cite{Wuebker12}.

To validate our experimental results, we measure the statistical significance using the paired bootstrap resampling method of \newcite{Koehn04}. Every result in this section is marked with $\ddag$ if it is statistically significantly better than the baseline with 95\% confidence, or with $\dag$ for 90\% confidence.

\subsection{Comparison of Vocabularies}
\label{sec:opt-wc}

The presented smoothing models are dependent on the label vocabulary, which is defined by the word-label mapping $c$. Here, we train the models with various label vocabularies and compare their smoothing performance.

The experiments are done on the IWSLT 2012 German$\rightarrow$English shared translation task. To rapidly perform repetitive experiments, we train the translation models with the in-domain TED portion of the dataset (roughly 2.5M running words for each side). We run the monolingual word clustering algorithm of \cite{Botros15} on each side of the parallel training data to obtain class label vocabularies (Section \ref{sec:wc}).

We carry out comparative experiments regarding the three factors of the clustering algorithm:

\begin{enumerate}[label=\textbf{\arabic*)}]
\item \textbf{Clustering iterations.} It is shown that the number of iterations is the most influential factor in clustering quality \cite{Och95}. We now verify its effect on translation quality when the clustering is used for phrase table smoothing.

As we run the clustering algorithm, we extract an intermediate class mapping for each iteration and train the smoothing models with it. The model weights are tuned for each iteration separately. The \textsc{Bleu} scores of the tuned systems are given in Figure \ref{fig:bleu-tune}. We use 100 classes on both source and target sides.

\begin{figure}[!ht]
\centering
\includegraphics[width=\linewidth]{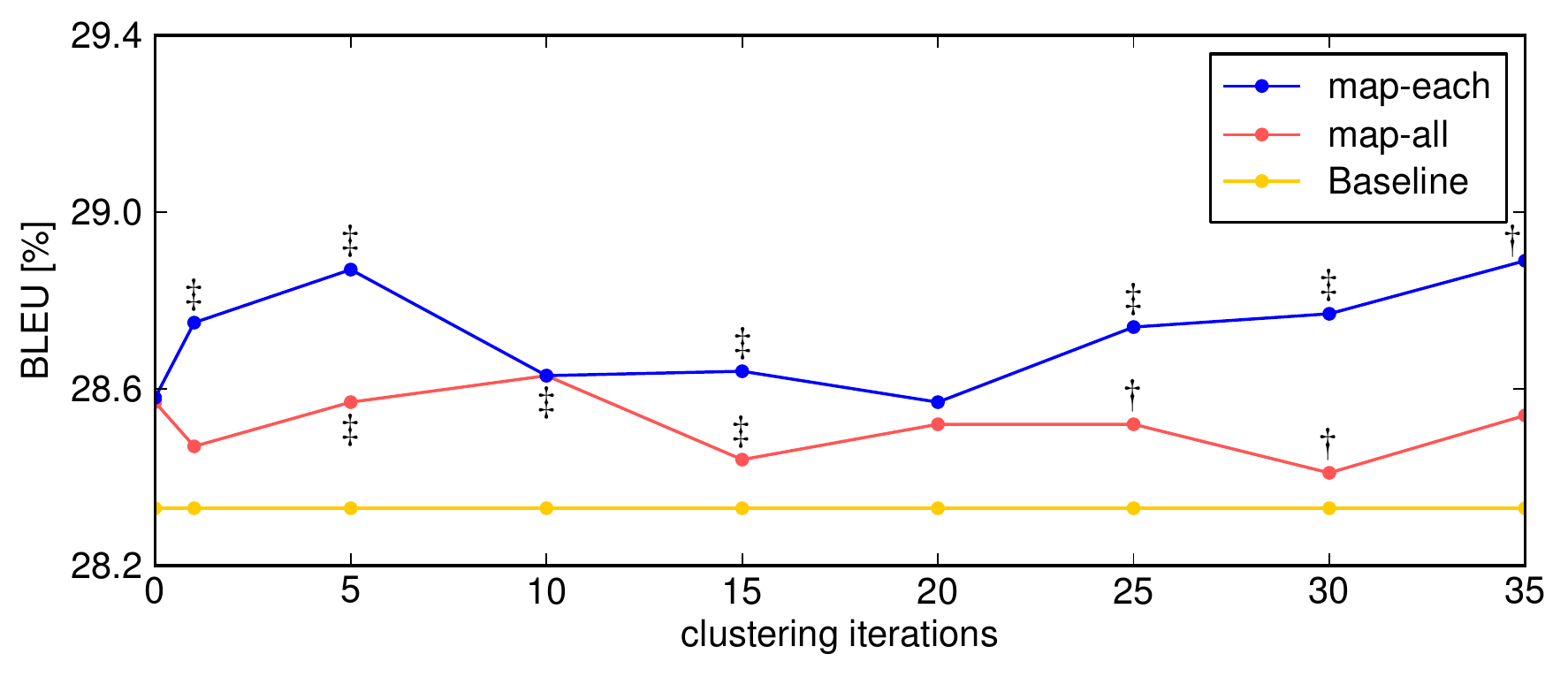}
\caption{\textsc{Bleu} scores for clustering iterations when using individually tuned model weights for each iteration. Dots indicate those iterations in which the translation is performed.}
\label{fig:bleu-tune}
\end{figure}

The score does not consistently increase or decrease over the iterations; it is rather on a similar level ($\pm\:0.2\%$ \textsc{Bleu}) for all settings with slight fluctuations. This is an important clue that the whole process of word clustering has no meaning in smoothing phrase translation models.

To see this more clearly, we keep the model weights fixed over different systems and run the same set of experiments. In this way, we focus only on the change of label vocabulary, removing the impact of nondeterministic model weight optimization. The results are given in Figure \ref{fig:bleu-lfix}.

\begin{figure}[!ht]
\centering
\includegraphics[width=\linewidth]{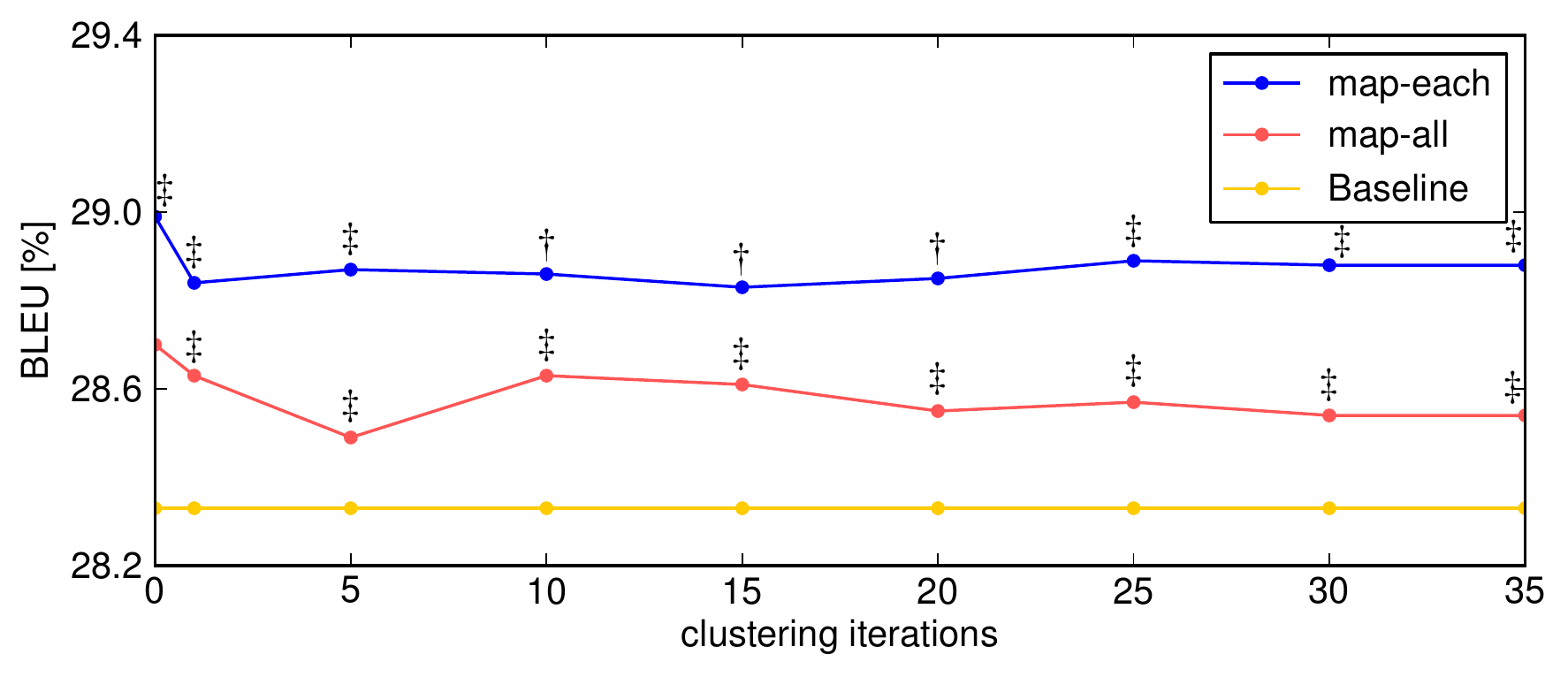}
\caption{\textsc{Bleu} scores for clustering iterations when using a fixed set of model weights. The weights that produce the best results in Figure \ref{fig:bleu-tune} are chosen.}
\label{fig:bleu-lfix}
\end{figure}

This time, the curves are even flatter, resulting in only $\pm\:0.1\%$ \textsc{Bleu} difference over the iterations. More surprisingly, the models trained with the initial clustering, i.e. when the clustering algorithm has not even started yet, are on a par with those trained with more optimized classes in terms of translation quality.

\item \textbf{Initialization of the clustering.} Since the clustering process has no significant impact on the translation quality, we hypothesize that the initialization may dominate the clustering. We compare five different initial class mappings:

\begin{itemize}
\item random: randomly assign words to classes
\item top-frequent (default): top-frequent words have their own classes, while all other words are in the last class
\item same-countsum: each class has almost the same sum of word unigram counts
\item same-\#words: each class has almost the same number of words
\item count-bins: each class represents a bin of the total count range
\end{itemize}

\begin{table}[!ht]
  \setlength{\tabcolsep}{5pt}
  \centering
	\begin{tabular}{lccc}
    \toprule
	 & & \textsc{Bleu} & \textsc{Ter}\\
   & Initialization & [\%] & [\%]\\
    \midrule
    Baseline & & 28.3 & 52.2\\
	\midrule
	+ map-each & random & \:\,28.9$^\ddag$ & \:\,51.7$^\ddag$\\
	 & top-frequent & \:\,29.0$^\ddag$ & \:\,51.5$^\ddag$\\
	 & same-countsum & \:\,28.8$^\ddag$ & \:\,51.7$^\ddag$\\
	 & same-\#words & \:\,28.9$^\ddag$ & \:\,51.6$^\ddag$\\
	 & count-bins & \:\,29.0$^\ddag$ & \:\,51.4$^\ddag$\\
	\bottomrule
  \end{tabular}
\caption{Translation results for various initializations of the clustering. 100 classes on both sides.}
\label{tab:initial}
\end{table}

Table \ref{tab:initial} shows the translation results with the map-each model trained with these initializations---without running the clustering algorithm. We use the same set of model weights used in Figure \ref{fig:bleu-lfix}. We find that the initialization method also does not affect the translation performance. As an extreme case, random clustering is also a fine candidate for training the map-each model.

\item \textbf{Number of classes.} This determines the vocabulary size of a label space, which eventually adjusts the smoothing degree. Table \ref{tab:numclasses} shows the translation performance of the map-each model with a varying number of classes. Similarly as before, there is no serious performance gap among different word classes, and POS tags and lemmas also comform to this trend.

However, we observe a slight but steady degradation of translation quality ($\approx\,$-0.2\% \textsc{Bleu}) when the vocabulary size is larger than a few hundreds. We also lose statistical significance for \textsc{Bleu} in these cases. The reason could be: If the label space becomes larger, it gets closer to the original vocabulary and therefore the smoothing model provides less additional information to add to the standard phrase translation model.

\begin{table}[!ht]
  \setlength{\tabcolsep}{4pt}
  \centering
	\begin{tabular}{lccc}
    \toprule
	& \#vocab & \textsc{Bleu} & \textsc{Ter}\\
  & (source) & [\%] & [\%]\\
    \midrule
	Baseline & & 28.3 & 52.2\\
	\midrule
	+ map-each & \enspace\enspace100 & \:\,29.0$^\ddag$ & \:\,51.5$^\ddag$\\
   (word class)  & \enspace\enspace200 & \:\,28.9$^\dag$ & \:\,51.6$^\ddag$\\
	 & \enspace\enspace500 & 28.7 & \:\,51.8$^\ddag$\\
	 & \enspace1000 & 28.7 & \:\,51.8$^\ddag$\\
	 & 10000 & 28.7 & \:\,51.9$^\dag$\\
	\midrule
	+ map-each (POS) & \enspace\enspace\enspace52 & \:\,28.9$^\dag$ & \:\,51.5$^\ddag$\\
	+ map-each (lemma) & 26744 & 28.8 & \:\,51.7$^\ddag$\\
	\bottomrule
  \end{tabular}
\caption{Translation results for different vocabulary sizes.}
\label{tab:numclasses}
\end{table}
\end{enumerate}

The series of experiments show that the map-each model performs very similar across vocabulary size and its structure. From our internal experiments, this argument also holds for the map-all model. The results do not change even when we use a different clustering algorithm, e.g. bilingual clustering \cite{Och99}. For the translation performance, the more important factor is the log-linear model training to find an optimal set of weights for the smoothing models.

\subsection{Comparison of Smoothing Models}
\label{sec:results}

\begin{table*}[!ht]
\centering
    \begin{tabular}{lcccccccc}
    \toprule
     & \multicolumn{2}{c}{IWSLT 2012} & \multicolumn{2}{c}{WMT 2015} & \multicolumn{2}{c}{WMT 2014} & \multicolumn{2}{c}{WMT 2015}\\
     & German & English & Finnish & English & English & German & English & Czech\\
    \midrule
	Sentences & \multicolumn{2}{c}{130k} & \multicolumn{2}{c}{1.1M} & \multicolumn{2}{c}{4M} & \multicolumn{2}{c}{0.9M}\\
	Running Words & 2.5M & 2.5M & 23M & 32M & 104M & 105M & 23.9M & 21M\\
	Vocabulary & 71k & 49k & 509k & 88k & 648k & 659k & 161k & 345k\\
	\bottomrule
    \end{tabular}
\caption{Bilingual training data statistics for IWSLT 2012 German$\rightarrow$English, WMT 2015 Finnish$\rightarrow$English, WMT 2014 English$\rightarrow$German, and WMT 2015 English$\rightarrow$Czech tasks.}
\label{tab:corpus-stat}
\end{table*}

\begin{table*}[!ht]
  \centering
  \begin{tabular}{lcccccccc}
    \toprule
     & \multicolumn{2}{c}{de-en} & \multicolumn{2}{c}{fi-en} & \multicolumn{2}{c}{en-de} & \multicolumn{2}{c}{en-cs}\\
     & \textsc{Bleu} & \textsc{Ter} & \textsc{Bleu} & \textsc{Ter} & \textsc{Bleu} & \textsc{Ter} & \textsc{Bleu} & \textsc{Ter}\\
     & [\%] & [\%] & [\%] & [\%] & [\%] & [\%] & [\%] & [\%]\\
    \midrule
    Baseline & 28.3 & 52.2 & 15.1 & 72.6 & 14.6 & 69.8 & 15.3 & 68.7\\
    \midrule
    + map-all & \:\,28.6$^\ddag$ & \:\,51.6$^\ddag$ & \:\,15.3$^\ddag$ & 72.5 & \:\,14.8$^\ddag$ & \:\,69.4$^\ddag$ & \:\,15.4$^\ddag$ & \:\,68.2$^\ddag$\\
	+ map-each & \:\,\textbf{29.0}$^\ddag$ & \:\,\textbf{51.4}$^\ddag$ & \:\,\textbf{15.8}$^\ddag$ & \:\,\textbf{72.0}$^\ddag$ & \:\,\textbf{15.1}$^\ddag$ & \:\,\textbf{69.0}$^\ddag$ & \:\,\textbf{15.8}$^\ddag$ & \:\,\textbf{67.6}$^\ddag$\\
	\bottomrule
  \end{tabular}
\caption{Translation results for IWSLT 2012 German$\rightarrow$English, WMT 2015 Finnish$\rightarrow$English, WMT 2014 English$\rightarrow$German, and WMT 2015 English$\rightarrow$Czech tasks.}
\label{tab:results}
\end{table*}

Next, we compare the two smoothing models by their performance in four different translation tasks: IWSLT 2012 German$\rightarrow$English, WMT 2015 Finnish$\rightarrow$English, WMT 2014 English$\rightarrow$German, and WMT 2015 English$\rightarrow$Czech. We train 100 classes on each side with 30 clustering iterations starting from the default (top-frequent) initialization.

Table \ref{tab:corpus-stat} provides the corpus statistics of all datasets used. Note that a morphologically rich language is on the source side for the first two tasks, and on the target side for the last two tasks. According to the results (Table \ref{tab:results}), the map-each model, which encourages backing off infrequent words, performs consistently better (maximum +0.5\% \textsc{Bleu}, -0.6\% \textsc{Ter}) than the map-all model in all cases.

\subsection{Comparison of Training Data Size}

\begin{figure}[!ht]
\centering
\includegraphics[width=\linewidth]{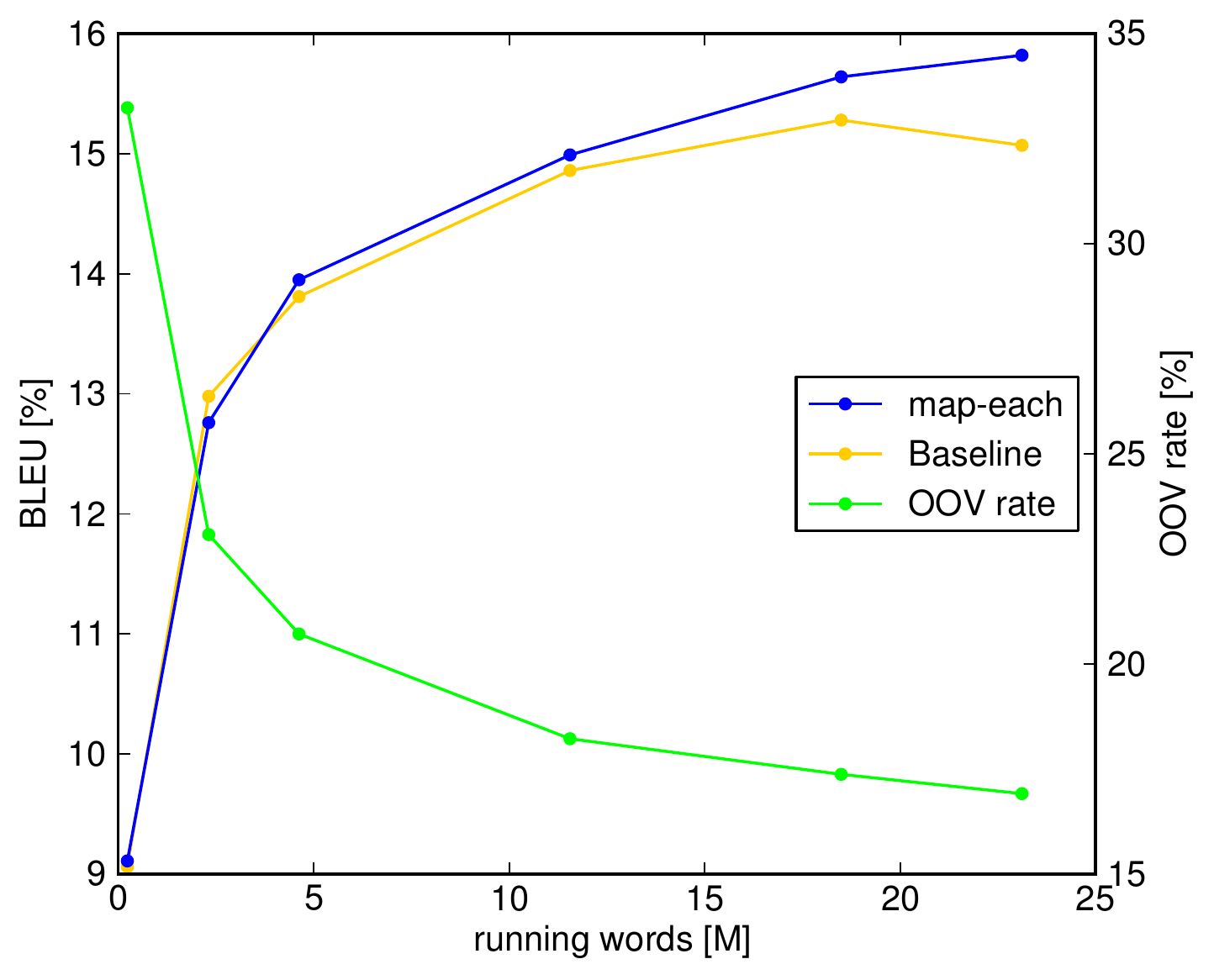}
\caption{\textsc{Bleu} scores and OOV rates for the varying training data portion of WMT 2015 Finnish$\rightarrow$English data.}
\label{fig:bleu-data}
\end{figure}

Lastly, we analyze the smoothing performance for different training data sizes (Figure \ref{fig:bleu-data}). The improvement of \textsc{Bleu} score over the baseline decreases drastically when the training data get smaller. We argue that this is because the smoothing models are only the additional scores for the phrases seen in the training data. For smaller training data, we have more out-of-vocabulary (OOV) words in the test set, which cannot be handled by the presented models.

\section{Analysis}
\label{sec:analysis}

\begin{table*}[!ht]
 \centering
  \begin{tabular}{ccccc}
    \toprule
    & & & \multicolumn{2}{c}{Top 200 \textsc{Ter}-improved Sentences}\\
    \cmidrule{4-5}
    & & & Common Input & Same Translation \\
    Model & Classes & \#vocab & [\%] & [\%]\\
    \midrule
    map-each & optimized & \enspace\enspace100 & - & -\\
    & non-optimized & \enspace\enspace100 & 89.5 & 89.9\\
    & random & \enspace\enspace100 & 88.5 & 89.8\\
	  & lemma & 26744 & 87.0 & 92.6\\
    \midrule
    map-all & optimized & \enspace\enspace100 & 56.0 & 54.5\\
    \bottomrule
  \end{tabular}
\caption{Comparison of translation outputs for the smoothing models with different vocabularies. ``optimized'' denotes 30 iterations of the clustering algorithm, whereas ``non-optimized'' means the initial (default) clustering.}
\label{tab:analysis}
\end{table*}

In Section \ref{sec:opt-wc}, we have shown experimentally that more optimized or more fine-grained classes do not guarantee better smoothing performance. We now verify by examining translation outputs that the same level of performance is not by chance but due to similar hypothesis scoring across different systems.

Given a test set, we compare its translations generated from different systems as follows. First, for each translated set, we sort the sentences by how much the sentence-level \textsc{Ter} is improved over the baseline translation. Then, we select the top 200 sentences from this sorted list, which represent the main contribution to the decrease of \textsc{Ter}. In Table \ref{tab:analysis}, we compare the top 200 $\textsc{Ter}$-improved translations of the map-each model setups with different vocabularies.

In the fourth column, we trace the input sentences that are translated by the top 200 lists, and count how many of those inputs are overlapped across given systems. Here, a large overlap indicates that two systems are particularly effective in a large common part of the test set, showing that they behaved analogously in the search process. The numbers in this column are computed against the map-each model setup trained with 100 optimized word classes (first row). For all map-each settings, the overlap is very large---around 90\%.

To investigate further, we count how often the two translations of a single input are identical (the last column). This is normalized by the number of common input sentences in the top 200 lists between two systems. It is a straightforward measure to see if two systems discriminate translation hypotheses in a similar manner. Remarkably, all systems equipped with the map-each model produce exactly the same translations for the most part of the top 200 $\textsc{Ter}$-improved sentences.

We can see from this analysis that, even though a smoothing model is trained with essentially different vocabularies, it helps the translation process in basically the same manner. For comparison, we also compute the measures for a map-all model, which are far behind the high similarity among the map-each models. Indeed, for smoothing phrase translation models, changing the model structure for vocabulary reduction exerts a strong influence in the hypothesis scoring, yet changing the vocabulary does not.

\section{Conclusion}
\label{sec:conc}

Reducing vocabulary using word-label mapping is a simple and effective way of smoothing phrase translation models. By mapping each word in a phrase at a time, the translation quality can be improved by up to +0.7\% \textsc{Bleu} and -0.8\% \textsc{Ter} over a standard phrase-based SMT baseline, which is superior to \newcite{Wuebker13}.

Our extensive comparison among various vocabularies shows that different word-label mappings are almost equally effective for smoothing phrase translation models. This allows us to use any type of word-level label, e.g. a randomized vocabulary, for the smoothing, which saves a considerable amount of effort in optimizing the structure and granularity of the label vocabulary. Our analysis on sentence-level \textsc{Ter} demonstrates that the same level of performance stems from the analogous hypothesis scoring.

We claim that this result emphasizes the fundamental sparsity of the standard phrase translation model. Too many target phrase candidates are originally undervalued, so giving them any reasonable amount of extra probability mass, e.g. by smoothing with random classes, is enough to broaden the search space and improve translation quality. Even if we change a single parameter in estimating the label space, it does not have a significant effect on scoring hypotheses, where many other models than the smoothed translation model, e.g. language models, are involved with large weights. Nevertheless, an exact linguistic explanation is still to be discovered.

Our results on varying training data show that vocabulary reduction is more suitable for large-scale translation setups. This implies that OOV handling is more crucial than smoothing phrase translation models for low-resource translation tasks.

For future work, we plan to perform a similar set of comparative experiments on neural machine translation systems.

\section*{Acknowledgments}

This paper has received funding from the European Union's Horizon 2020 research and innovation programme under grant agreement n\textsuperscript{o}~645452 (QT21).

\bibliographystyle{acl2016}
\bibliography{references}

\begin{thebibliography}{}

\bibitem[\protect\citename{Bisazza and Monz}2014]{Bisazza14}
Arianna Bisazza and Christof Monz.
\newblock 2014.
\newblock Class-based language modeling for translating into morphologically
  rich languages.
\newblock In {\em Proceedings of 25th International Conference on Computational
  Linguistics (COLING 2014)}, pages 1918--1927, Dublin, Ireland, August.

\bibitem[\protect\citename{Botros \bgroup et al.\egroup }2015]{Botros15}
Rami Botros, Kazuki Irie, Martin Sundermeyer, and Hermann Ney.
\newblock 2015.
\newblock On efficient training of word classes and their application to
  recurrent neural network language models.
\newblock In {\em Proceedings of 16th Annual Conference of the International
  Speech Communication Association (Interspeech 2015)}, pages 1443--1447,
  Dresden, Germany, September.

\bibitem[\protect\citename{Brown \bgroup et al.\egroup }1992]{Brown92}
Peter~F. Brown, Peter~V. deSouza, Robert~L. Mercer, Vincent J.~Della Pietra,
  and Jenifer~C. Lai.
\newblock 1992.
\newblock Class-based n-gram models of natural language.
\newblock {\em Computational Linguistics}, 18(4):467--479, December.

\bibitem[\protect\citename{Cherry}2013]{Cherry13}
Colin Cherry.
\newblock 2013.
\newblock Improved reordering for phrase-based translation using sparse
  features.
\newblock In {\em Proceedings of 2013 Conference of the North American Chapter
  of the Association for Computational Linguistics: Human Language Technologies
  (NAACL-HLT 2013)}, pages 22--31, Atlanta, GA, USA, June.

\bibitem[\protect\citename{Durrani \bgroup et al.\egroup }2014]{Durrani14}
Nadir Durrani, Philipp Koehn, Helmut Schmid, and Alexander Fraser.
\newblock 2014.
\newblock Investigating the usefulness of generalized word representations in
  smt.
\newblock In {\em Proceedings of 25th Annual Conference on Computational
  Linguistics (COLING 2014)}, pages 421--432, Dublin, Ireland, August.

\bibitem[\protect\citename{Galley and Manning}2008]{Galley08}
Michel Galley and Christopher~D. Manning.
\newblock 2008.
\newblock A simple and effective hierarchical phrase reordering model.
\newblock In {\em Proceedings of 2008 Conference on Empirical Methods in
  Natural Language Processing (EMNLP 2008)}, pages 848--856, Honolulu, HI, USA,
  October.

\bibitem[\protect\citename{Haddow \bgroup et al.\egroup }2015]{Haddow15}
Barry Haddow, Matthias Huck, Alexandra Birch, Nikolay Bogoychev, and Philipp
  Koehn.
\newblock 2015.
\newblock The edinburgh/jhu phrase-based machine translation systems for wmt
  2015.
\newblock In {\em Proceedings of 2016 EMNLP 10th Workshop on Statistical
  Machine Translation (WMT 2016)}, pages 126--133, Lisbon, Portugal, September.

\bibitem[\protect\citename{Koehn and Hoang}2007]{Koehn07}
Philipp Koehn and Hieu Hoang.
\newblock 2007.
\newblock Factored translation models.
\newblock In {\em Proceedings of 2007 Joint Conference on Empirical Methods in
  Natural Language Processing and Computational Natural Language Learning
  (EMNLP-CoNLL 2007)}, pages 868--876, Prague, Czech Republic, June.

\bibitem[\protect\citename{Koehn \bgroup et al.\egroup }2003]{Koehn03}
Philipp Koehn, Franz~Josef Och, and Daniel Marcu.
\newblock 2003.
\newblock Statistical phrase-based translation.
\newblock In {\em Proceedings of 2003 Conference of the North American Chapter
  of the Association for Computational Linguistics on Human Language Technology
  (NAACL-HLT 2003)}, pages 48--54, Edmonton, Canada, May.

\bibitem[\protect\citename{Koehn}2004]{Koehn04}
Philipp Koehn.
\newblock 2004.
\newblock Statistical significance tests for machine translation evaluation.
\newblock In {\em Proceedings of 2004 Conference on Empirical Methods in
  Natural Language Processing (EMNLP 2004)}, pages 388--395, Barcelona, Spain,
  July.

\bibitem[\protect\citename{Koehn}2010]{Koehn10}
Philipp Koehn.
\newblock 2010.
\newblock {\em Statistical Machine Translation}.
\newblock Cambridge University Press, New York, NY, USA.

\bibitem[\protect\citename{Koo \bgroup et al.\egroup }2008]{Koo08}
Terry Koo, Xavier Carreras, and Michael Collins.
\newblock 2008.
\newblock Simple semi-supervised dependency parsing.
\newblock In {\em Proceedings of 46th Annual Meeting of the Association for
  Computational Linguistics (ACL 2008)}, pages 595--603, Columbus, OH, USA,
  June.

\bibitem[\protect\citename{Miller \bgroup et al.\egroup }2004]{Miller04}
Scott Miller, Jethran Guinness, and Alex Zamanian.
\newblock 2004.
\newblock Name tagging with word clusters and discriminative training.
\newblock In {\em Proceedings of 2004 Conference of the North American Chapter
  of the Association for Computational Linguistics: Human Language Technologies
  (NAACL-HLT 2004)}, pages 337--342, Boston, MA, USA, May.

\bibitem[\protect\citename{Och}1995]{Och95}
Franz~Josef Och.
\newblock 1995.
\newblock Maximum-likelihood-sch\"{a}tzung von wortkategorien mit verfahren der
  kombinatorischen optimierung.
\newblock Studienarbeit, Friedrich-Alexander-Universit\"{a}t
  Erlangen-N\"{u}rnberg, Erlangen, Germany, May.

\bibitem[\protect\citename{Och}1999]{Och99}
Franz~Josef Och.
\newblock 1999.
\newblock An efficient method for determining bilingual word classes.
\newblock In {\em Proceedings of 9th Conference on European Chapter of
  Association for Computational Linguistics (EACL 1999)}, pages 71--76, Bergen,
  Norway, June.

\bibitem[\protect\citename{Och}2003]{Och03mert}
Franz~Josef Och.
\newblock 2003.
\newblock Minimum error rate training in statistical machine translation.
\newblock In {\em Proceedings of 41st Annual Meeting of the Association for
  Computational Linguistics (ACL 2003)}, pages 160--167, Sapporo, Japan, July.

\bibitem[\protect\citename{Rish{\o}j and S{\o}gaard}2011]{Rishoj11}
Christian Rish{\o}j and Anders S{\o}gaard.
\newblock 2011.
\newblock Factored translation with unsupervised word clusters.
\newblock In {\em Proceedings of 2011 EMNLP 6th Workshop on Statistical Machine
  Translation (WMT 2011)}, pages 447--451, Edinburgh, Scotland, July.

\bibitem[\protect\citename{Wuebker \bgroup et al.\egroup }2012]{Wuebker12}
Joern Wuebker, Matthias Huck, Stephan Peitz, Malte Nuhn, Markus Freitag,
  Jan-Thorsten Peter, Saab Mansour, and Hermann Ney.
\newblock 2012.
\newblock Jane 2: Open source phrase-based and hierarchical statistical machine
  translation.
\newblock In {\em Proceedings of 24th International Conference on Computational
  Linguistics (COLING 2012)}, pages 483--492, Mumbai, India, December.

\bibitem[\protect\citename{Wuebker \bgroup et al.\egroup }2013]{Wuebker13}
Joern Wuebker, Stephan Peitz, Felix Rietig, and Hermann Ney.
\newblock 2013.
\newblock Improving statistical machine translation with word class models.
\newblock In {\em Proceedings of 2013 Conference on Empirical Methods in
  Natural Language Processing (EMNLP 2013)}, pages 1377--1381, Seattle, USA,
  October.

\bibitem[\protect\citename{Zens \bgroup et al.\egroup }2002]{Zens02}
Richard Zens, Franz~Josef Och, and Hermann Ney.
\newblock 2002.
\newblock Phrase-based statistical machine translation.
\newblock In Matthias Jarke, Jana Koehler, and Gerhard Lakemeyer, editors, {\em
  25th German Conference on Artificial Intelligence (KI2002)}, volume 2479 of
  {\em Lecture Notes in Artificial Intelligence (LNAI)}, pages 18--32, Aachen,
  Germany, September. Springer Verlag.

\end{thebibliography}

\end{document}